\documentclass[letterpaper, 10 pt, conference]{ieeeconf}  
\IEEEoverridecommandlockouts

\usepackage{cite}
\usepackage{amsmath,amssymb,amsfonts}
\usepackage{algorithmic}
\usepackage{enumerate}
\usepackage{graphicx}
\usepackage{hyperref}
\usepackage{textcomp}
\usepackage{xcolor}
\usepackage{array}
\usepackage{mwe}
\usepackage{caption,subcaption}

\hypersetup{
    colorlinks=true,
    linkcolor=blue,
    filecolor=magenta,      
    urlcolor=cyan,
}

\newcommand{\Softbubble}{\emph{Soft-bubble} } 
\newcommand{\Softbubbles}{\emph{Soft-bubbles} }

\newif\ifcomments
\commentstrue 

\def\BibTeX{{\rm B\kern-.05em{\sc i\kern-.025em b}\kern-.08em
    T\kern-.1667em\lower.7ex\hbox{E}\kern-.125emX}}
    
\title{Variable compliance and geometry regulation of Soft-Bubble grippers with active pressure control}
\author{Sihah Joonhigh*, Naveen Kuppuswamy*, Andrew Beaulieu*, Alex Alspach, and Russ Tedrake \\
\thanks{*Authors contributed equally.}
  Toyota Research Institute (TRI) \\
  \texttt{first.lastname@tri.global} \\}

\begin{document}

\maketitle

\begin{abstract}

While compliant grippers have become increasingly commonplace in robot manipulation, finding the right stiffness and geometry for grasping the widest variety of objects remains a key challenge. Adjusting both stiffness and gripper geometry on the fly may provide the versatility needed to manipulate the large range of objects found in domestic environments. We present a system for actively controlling the geometry (inflation level) and compliance of Soft-bubble grippers - air filled, highly compliant parallel gripper fingers incorporating visuotactile sensing. The proposed system enables large, controlled changes in gripper finger geometry and grasp stiffness, as well as simple in-hand manipulation. We also demonstrate, despite these changes, the continued viability of advanced perception capabilities such as dense geometry and shear force measurement - we present a straightforward extension of our previously presented approach for measuring shear induced displacements using the internal imaging sensor and taking into account pressure and geometry changes. We quantify the controlled variation of grasp-free geometry, grasp stiffness and contact patch geometry resulting from pressure regulation and we demonstrate new capabilities for the gripper in the home by grasping in constrained spaces, manipulating tools requiring lower and higher stiffness grasps, as well as contact patch modulation.
\end{abstract}

\section{Introduction}
The challenge of precisely grasping objects under pose and shape uncertainties has driven the development of compliant and sensitive gripper systems that have now become a commonplace component of manipulation hardware, particularly in the case of robots built for our homes. Regardless of a gripper's simplicity or complexity, equipping the contact surfaces with the right amount of passive compliance has proved to be tremendously useful in facilitating stable, reliable grasps \cite{Hughes2016}. Our homes are complex, cluttered and constrained and the objects we have vary widely in shape and size. This environmental complexity typically inspires more complexity in the gripper by way of tactile sensing, control strategies and more degrees of freedom (DOF) that enable a gripper to adapt to the task at hand, e.g., sliding fingertips under a plate, stiffening and relaxing or opening wide to grasp a larger object. While some robots employ advanced high-DOF grippers with the ability to change gripping geometry and stiffness, many fall back on less capable single-DOF grippers for their strength and reliability. We aim to enable finger geometry and stiffness control on a simple parallel jaw gripper with \emph{Soft-bubble} fingers by actively controlling the internal pressures of each bubble.

\begin{figure}[t]
    \centering 
    \includegraphics[width=0.95\columnwidth]{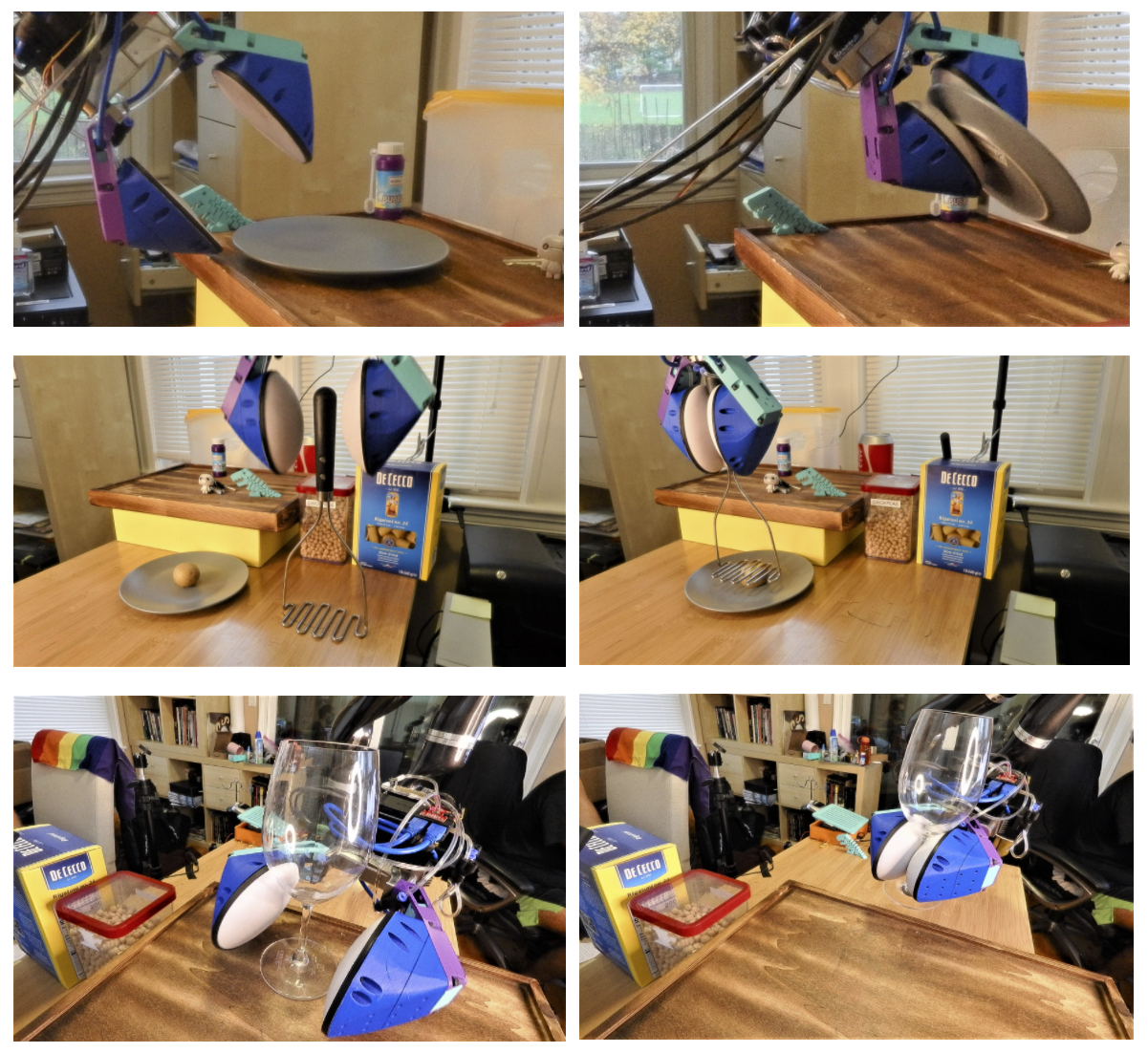}
    \caption{The utility of variable geometry and compliance regulation in domestic tasks: Geometry change of the \Softbubble grippers in form of deflation enables sliding the gripper into tight spaces such as between a plate and a table top (top-left), and re-inflation enables a stable grasp on a narrow, flat form factor object (top-right). Compliance regulation enables stable grasps on tools (middle-left) while increased stiffness from increased pressure enables high-stiffness actions like mashing a potato (middle-right). Lastly, geometry change from pressure reduction enables positioning the gripper around objects such as the wine-glass stem even under pose variance (bottom-left) and re-inflation generates stable grasps (bottom-right).}
    \label{fig:Geometry_compliance_control}
\end{figure}

While compliant grippers have been an active research focus in manipulation, novel materials and manufacturing techniques have increasingly driven the designs towards soft materials \cite{shintake2018soft} which can be fabricated in the desired geometries and compliance. While variable compliance has a long history in robotics \cite{wolf2015variable}, particularly in the domain of legged locomotion, applications in manipulation, and in grasping in particular, have been less common. The design of effective variable compliance mechanisms for grippers can be a complex challenge for following reasons: (i) The relatively small desired form factor of grippers imposes size restrictions on mechanisms, (ii) mechanism weight is limited due to payload limitations of robot arms, and (iii) the need for mechanical robustness against unexpected contacts and collisions. Additional challenges may also arise from the requirement to incorporate into the grippers some form of tactile sensing in order to sense the state of the gripper or the object being manipulated.

Unlike variable compliance, the need for variable geometry regulation is more nuanced; some applications for varying the geometry include grasping in tightly constrained spaces or for facilitating non-prehensile manipulation actions such as pushing objects \cite{ruggiero2018nonprehensile}. In the case of soft grippers, geometry change enables the positioning of grippers within tightly constrained spaces, such as in cluttered home environments, as well as the positioning of grippers effectively against complex geometries such as within or around handles of mugs or between plates. In any case, incorporating the grippers with the right kind of proprioception in order to regulate the geometry change is vital since inferring the current shape without any geometric feedback could be computationally intractable due to the nature of the governing continuum mechanics equations \cite{kuppuswamy2019fast}.

The key requirements for an effective and reliable variable compliance and variable geometry gripping system are therefore that of mechanical simplicity, large range of achievable geometries and compliance, and the ability to integrate with some form of contact sensing. One possibility towards achieving these goals is to leverage air-filled robotic structures. In this context, \Softbubble grippers are a recently demonstrated compliant gripper \cite{Kuppuswamy2020} whose fingers include a depth camera-based tactile sensor ensconced in an extremely soft air-filled membrane structure \cite{Alspach2019}. The mechanism of the \Softbubble finger itself consists of no moving parts and the mechanical separation of the imaging sensor and the inflated structure enables seamless decoupling of the compliance from the tactile sensing. In previously published research, the choice of membrane material, geometry and internal pressure were made based on the requirement of achieving robust grasps on select domestic objects. The passive mechanics and friction of the gripper were exploited and grasp forces were regulated using only the underlying parallel gripper controller. The mechanical simplicity and the ability to pressurize the passively compliant gripping structures make it an ideal candidate for demonstrating variable compliance and geometry for more capable grasping.

In this paper, we propose an augmentation to the \Softbubble gripper design that explicitly leverages the air-filled nature of these bubbles by using active pressure control which, as we demonstrate through our results, can regulate the gripping surface-geometry, gripper-object slip and grasp stiffness. As demonstrated in prior work, as a tactile sensor, the \Softbubble captures three principal modalities: (i) depth maps, (ii) IR images of the elastic surface, and (iii) internal air pressure. Previously, the tracking of shear forces was enabled by tracking the motion of markers printed  on the elastic surface. In this paper we present an extension of the algorithm in \cite{Kuppuswamy2020} that takes into account the geometry changes and stretching that can result from active pressure regulation and combine a so-called contact-patch estimator with the shear estimates. 

The proposed approaches were demonstrated using a set of real-world home manipulation tasks: (i) Active geometry change for (a) flat form factor object (e.g., plate) picking,  (b) narrow form-factor object grasping (e.g., wine-glass stem), and (c) wide form-factor grasping (e.g., mug); (ii) Active contact friction regulation (e.g., mug reorientation); and (iii) Active stiffness control for forceful tool-use (e.g., potato-mashing). 

This paper is organized as follows: Section \ref{sec:background} presents the relevant background. The proposed pressure regulation setup is presented in Section \ref{sec:pressure_regulation}, followed by the algorithm for shear estimation under geometric variation in Section \ref{sec:shear_estimation}. The experiments conducted are described in detail in Section \ref{sec:experiments}, followed by the conclusions in Section \ref{sec:conclusions}.

\section{Background}
\label{sec:background}
Soft or compliant grippers have a long history in manipulation \cite{schmidt1978flexible} and are increasingly seen as an important component towards improving grasp robustness under geometric and/or pose uncertainty \cite{Hughes2016}. While achievement of the requisite grasp stability can drive the mechanism design and control algorithm development \cite{Chang2019}, exploiting the passive mechanical properties of novel soft materials remains a popular approach \cite{shintake2018soft}. Softness, in this case, plays an important role in achieving grasp robustness by enabling the gripping surface to comply to the geometry of the manipulated object and thus results in increased contact area and friction between the gripper and the grasped object.

Multi-DOF grippers are one possibility for compliant and variable geometry gripping \cite{ozawa2017grasp}, in particular, for dexterous manipulation. While geometry change can be achieved by simply controlling the joint-space configuration of the gripper, incorporation of some kind of impedance control and force regulation, either through finger-tip force/tactile sensing or through proprioceptive means, enables active regulation of grasp compliance. However, parallel grippers possess a much greater extent of mechanical robustness in comparison. 

While a variety of popular material technologies have been investigated toward creating soft skins for manipulators - elastomers (e.g., silicone, shape memory materials, active polymers and gels) being popular examples \cite{schmitt2018}, an extension of these approaches is to utilize air-filled membranes as the surfaces and end-effectors of rigid links \cite{Kim2015, Alspach2015}. This approach can also lead to the development of effective tactile sensing grippers and end-effectors by incorporating an internal imaging sensor \cite{Alspach2019}. A key advantage of such inflated systems is that regulating the pressure of the air within the elastomeric membrane varies the geometry and the stiffness of the contact surfaces. The employment of camera-based sensing within these sensors \cite{Yamaguchi2019} largely decouples the sensing concerns from the geometry/compliance regulation concerns, i.e., it is possible to make changes to the compliant geometry with little or no changes to sensing electronics. 

Additional examples of variable stiffness grippers \cite{arachchige2020novel} and impedance control in pneumatic actuators \cite{paoletti2017grasping} indicate the importance of pressure regulation for grasping. Several state of art methods for pressure regulation have been proposed \cite{adami2019board} while model-based stiffness and position control of inflatable soft robots \cite{Best2020} remains a popular theme in soft robotics, despite the emergence of deep learning. Ideally, these methods result in robust hardware and fast computability.

\subsection{Tactile shear estimation}
In previously demonstrated results, dense optic flow techniques \cite{Farneback2003} have been successfully employed for tracking external shear and slip in inflated camera-based tactile sensors \cite{Yuan2015Shear} through the tracking of internal surface visual markers from within the sensor itself \cite{Kuppuswamy2020}. 

In the case of \Softbubbles, an additional complexity introduced in the shear-estimation problem is due to the large deformations induced on the membrane surface during grasping. There are two key components to the resulting motion of the membrane when subject to tangential shear: (i) the area of the membrane in direct contact with the object (the so-called \emph{contact patch}) remains in stiction with the object and thus translates along with the object, and (ii) the area not in contact is simply subject to the resultant forces from tangential elastic stresses and normal (expansive) forces due to the higher internal air pressure \cite{Taylor2005_fem_implementation}. While the former is a direct measure of shear induced deformation, the latter is an indirect measure and arguably harder to interpret (and utilize in feedback) without employing inverse-FEM based methods.

The need to algorithmically distinguish the non-contacted and contacted regions on the surface of this kind of membrane-based tactile sensor motivated development of a computationally efficient inverse method \cite{kuppuswamy2019fast}; a key limitation of that work was that the model does not take into account shear force. Therefore, for this work we utilized a \emph{naive} contact patch estimator similar to that proposed in \cite{Kuppuswamy2020}. 

From a perception standpoint, in the case of camera-based tactile sensors \cite{Shimonomura2019}, state-of-art approaches for shear and slip measurement utilize some form of tracking of markers etched, cast or printed on a sensor's surface \cite{yuan2015measurement, Yamaguchi2019}. In the case of \Softbubbles, the large degree of deformation of the membrane surface necessitates the usage of non-uniform, dense marker patterns to cope with the non-uniform nature of the membrane stretching under contact \cite{Kuppuswamy2020}. Some common applications of shear sensing include detection of the onset of slip, for the control of minimum necessary grasp force in the case of fragile objects, or towards coping with expected or unexpected contact between the environment and the grasped objects. For controlling the stiffness of grippers, shear sensing becomes an increasingly important.

\section{Pressure regulation setup}
\label{sec:pressure_regulation}
\begin{figure}[t]
    \centering 
    \includegraphics[width=0.94\columnwidth]{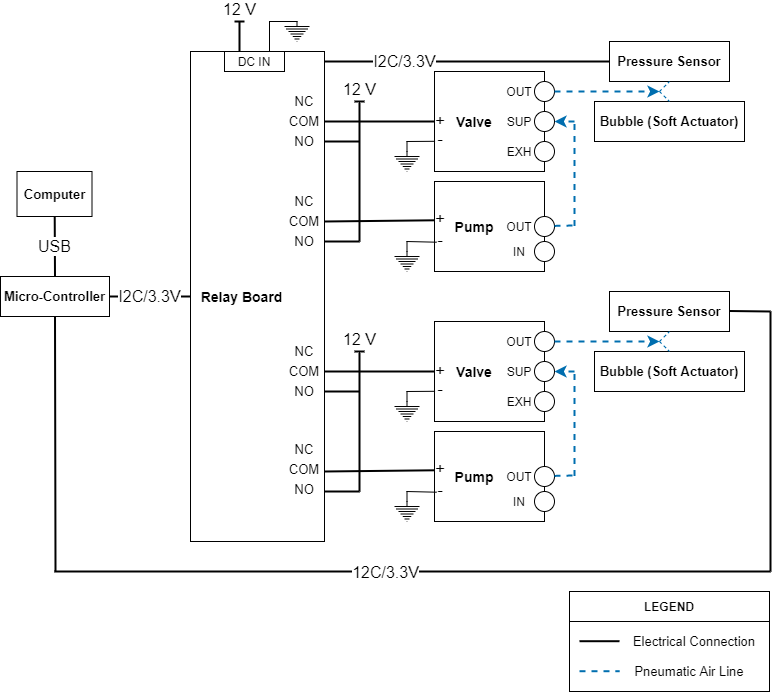}
    \caption{A schematic of the pressure regulation system for variable geometry and compliance regulation of air-filled sensors.}
    \label{fig:pressure_regulation_setup}
\end{figure}

The pressure regulation strategy we employ utilizes a compact rotary diaphragm pump (Koge Mini Air Pump, KPM32E-12A02), a three-port electronically controlled pneumatic valve (SMC 3 Port Solenoid Valve, S070C-6DG-32), an air pressure sensor (SparkFun Qwiic MicroPressure Sensor, SEN-16476), a relay board (Sparkfun Qwiic Quad Relay, COM-16566), an ARM-based microcontroller board (Sparkfun Redboard Artemis Nano, DEV-15443), 2~mm~ID~x~4~mm OD Clear Masterkleer Soft PVC Plastic Tubing, Y-connectors, and push-to-connect fittings with shut-off valves. These components create an inexpensive and compact pressure regulation setup that can be mounted at the robot end-effectors or elsewhere. The mass of the pressure regulation system is 450~g. 

The pressure regulation control system is implemented on a microcontroller which operates the pneumatic pump and solenoid exhaust valve via relays based on set points commanded via the robot control PC. We use a pressure sensor attached to the pneumatic lines in order to sense the air pressure within the system. Both the relay board and the pressure sensor that we used are controlled along the same I2C bus by the microcontroller. In our experiments, the pressure regulation system is located at the base of the robot. During the inflation and deflation processes, the pressure gradient that is created through the tubing that is run along the robot (approximately 2 meters) is non-negligible and must be factored into the control system. To minimize the effect that the pressure differential has on the measurement, the pressure sensor is placed as close along the pneumatic line to the \Softbubble (further from the pump and exhaust valve) as possible. For inflation and deflation this pressure differential is easily characterized and allows us to incorporate an offset correction to attain the correct resultant pressure in the \Softbubble.

The microcontroller receives serial commands sent through USB from the client computer in order to control the inflation of the \Softbubble. The client can set a target pressure and the microcontroller will inflate or deflate a bubble to reach the set point using the pump and exhaust valve, respectively. A pressure sensor for each independently controlled bubble provides feedback to the control system. 

\section{Shear Estimation under varying geometry}
\label{sec:shear_estimation}
This method computes a contact patch mask $M_k$ from a pair of depth images $D_0$ (reference depth image) and $D_k$ (current depth image); this mask can then be used for the computation of shear-induced displacement. 

\begin{figure}[t]
    \centering 
\begin{subfigure}{0.20\textwidth}
  \includegraphics[width=\linewidth]{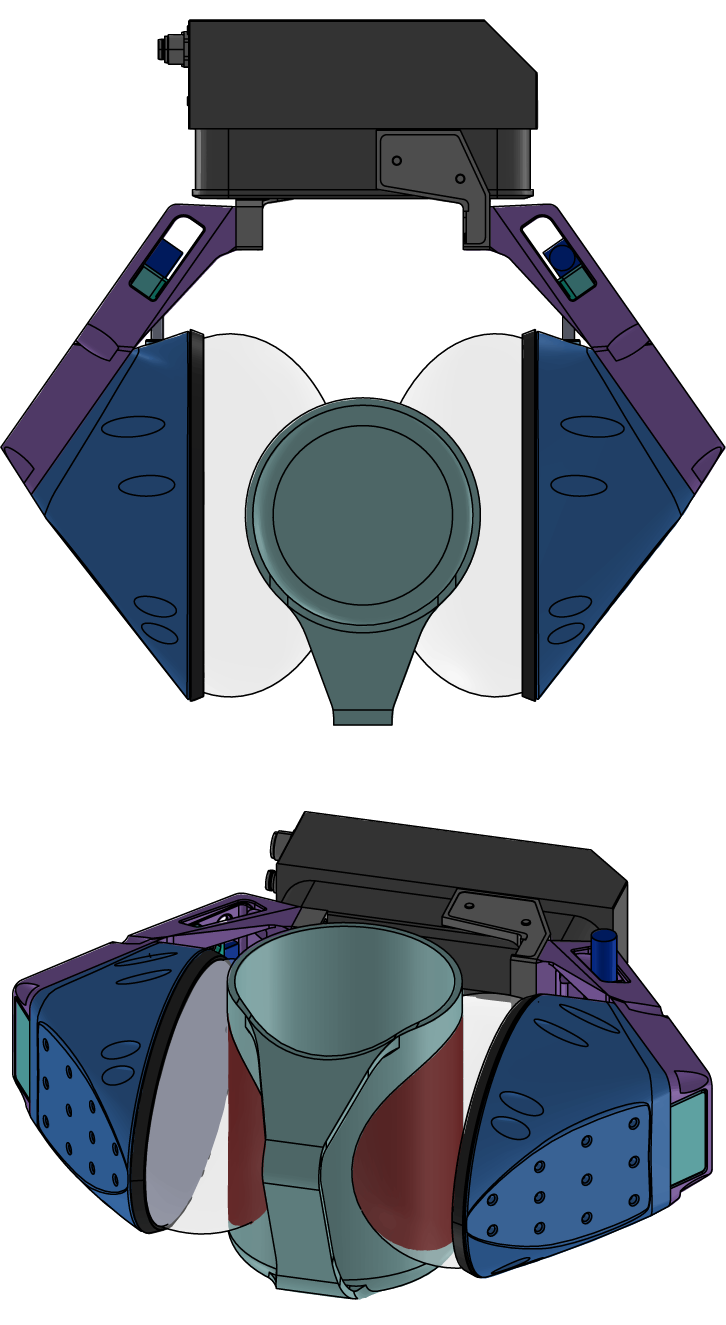}
  \caption{High Inflation}
  \label{fig:Geometry_compliance_control:high}
\end{subfigure}\hfil 
\begin{subfigure}{0.20\textwidth}
  \includegraphics[width=\linewidth]{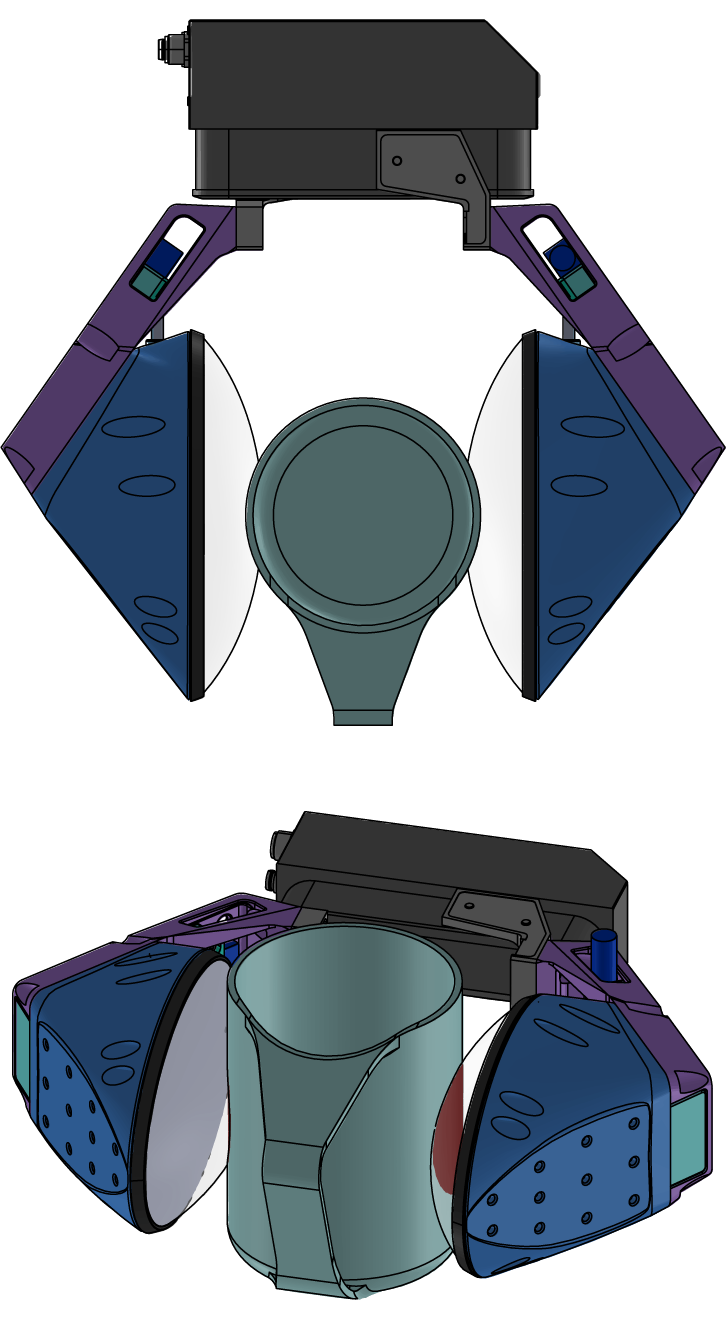}
  \caption{Low Inflation}
  \label{fig:Geometry_compliance_control:low}
\end{subfigure}\hfil 
\caption{A rendering of contact patch variation resulting from pressure and geometry changes while maintaining a set grasp width. Increased contact patch size leads to increased friction between the gripper and the object.}
\label{fig:need_for_contact_patch}
\end{figure}

Furthermore, under variable pressure conditions, it is important to understand how the reference depth image $D_0$ needs to be set. Fig. \ref{fig:need_for_contact_patch} illustrates the potential change in contact patch size resulting from a pressure change. While this rendered figure may not accurately capture the additional bulging of the non-contacted region outwards, the contact patch area clearly increases and thus results in a greater amount of friction between the grasped object and the gripper. We propose a solution that estimates the contact patch using a depth difference between non-contacted and contacted membrane geometries. Errors in capturing the boundaries of the contact patches have a negligible effect in the computation of the resultant shear-induced deformations in the optic flow as we are interested in relative, not absolute, measurements. This technique can be described as follows. 

Consider a pair of IR (grayscale) images $I_0$, $I_k$ obtained at two different grasp states - grasp state in this context could refer to conditional changes on the state of the grasped object - for instance, through contact mode changes due to an external contact, or grasp stiffness change due to pressure changes. 

Now, consider the binary mask image produced from naive contact patch estimation. This mask image $M_{0,k}$ is produced by taking a per-pixel depth difference between the depth images $D_0$ and $D_k$. This mask image can then be used to apply a threshold on the IR image $I_0$ and $I_k$ to generate the following masked IR images as,

\begin{equation}
\begin{split}
    \bar{I}_0 &= M_{0,k} * I_0 \\
    \bar{I}_k &= M_{0,k} * I_k.
\end{split}
\end{equation}

Using similar assumptions to other approaches \cite{Yuan2015Shear, Kuppuswamy2020}, we can compute the dense optic flow \cite{Farneback2003} purely on the masked IR images by,

\begin{equation}
    \bar{V}_{0 \rightarrow k} = f(\bar{I}_0, \bar{I}_k)
\end{equation}

The computation of the shear force from this optic flow image can then be performed by vector decomposition and summation techniques \cite{zhang2019} and then multiplied by an appropriate scaling gain matrix K to provide a three-dimensional estimate of net shear force on the contact patch. Fig. \ref{Shear_sensing} depicts the computed shear for the various experimental scenarios. The computational steps in the proposed shear computation algorithm are depicted in Fig. \ref{fig:pressure_regulation_setup}.

\begin{figure}[t]
    \centering 
    \includegraphics[width=1.0\columnwidth]{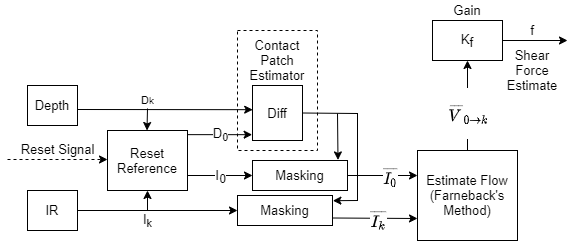}
    \caption{A schematic of the shear estimation algorithm. The effect of changing geometry and stiffness is taken into account through the utilization of a Reset Signal that can be triggered by high-level robot motion controllers based on grasp state. The symbols are described in Table \ref{table:symbols}. }
    \label{fig:pressure_regulation_setup}
\end{figure}

\begin{center}
\begin{tabular}{ |m{1cm} | m{6.5cm}|}
 \hline
 \textbf{Symbol} & \textbf{Explanation} \\ 
 \hline
 $D_i$ & A depth image captured by the internal imaging sensor at time i. \\ 
 $I_i$ & An IR image captured by the internal imaging sensor at time i.\\
 $M_{i,j}$ & A binary \emph{mask} image produced from the per-pixel depth differences between depth images $Di$ and $Dj$\\ 
 $\bar{I}_i$ & A masked IR image resulting from application of a mask $M_{i,j}$ onto an image $I_i$.\\ 
 $\bar{V}_{i \rightarrow j}$ & Dense optic flow computed between two images obtained at time $i$ and $j$ \\
 $K_f$ & Gain constant converting shear-induced displacement into a shear force estimate\\
 \hline
\end{tabular}
\captionof{table}{Explanation of the symbols used in the shear-estimation algorithm}
\label{table:symbols}
\end{center}

\section{Experiments}
\label{sec:experiments}
In order to quantify and demonstrate the utility of the proposed compliance and geometry regulation, we performed a set of home robotics tasks. The demonstrations as seen in Fig. \ref{fig:Geometry_compliance_control} were aimed at showing two kinds applications for the variable pressure enabled soft-bubble grippers: (i) varying pre-grasp geometry in order to fit the fingers into narrow form factor spaces - for eg. in tasks like plate picking, grasping wider form factor objects like mugs; and (ii) varying grasp stiffness - for eg. in forceful tool-use with a kitchen tool.

\subsection{Experimental Setup}
The setup consists of a pair of \emph{Soft-bubble} sensors mounted on a Schunk WSG-32 gripper. This gripper is attached to the end of a Kinova Jaco Gen2 robot arm which is mounted upright at the end of workbench. The design of the \Softbubble gripper was slightly modified from previously reported versions \cite{Kuppuswamy2020} to utilize the COTS version of the Picoflexx PMDTech ToF camera. The gripper has a max width of 66~mm. The bubbles are positioned in a manner as to barely make contact under standard inflation (1050~hPa) when the gripper is fully closed (0~mm commanded width). The robot planning and control was developed using the Drake simulation and control library \cite{bib:drake}. For prototyping the behaviors demonstrated in this paper, no external perception was used and the actions such as grasping, moving forward in Cartesian space until touch, etc., utilize force feedback but were artisinally crafted. 

\begin{figure}[ht]
    \centering 
    \includegraphics[width=0.95\columnwidth]{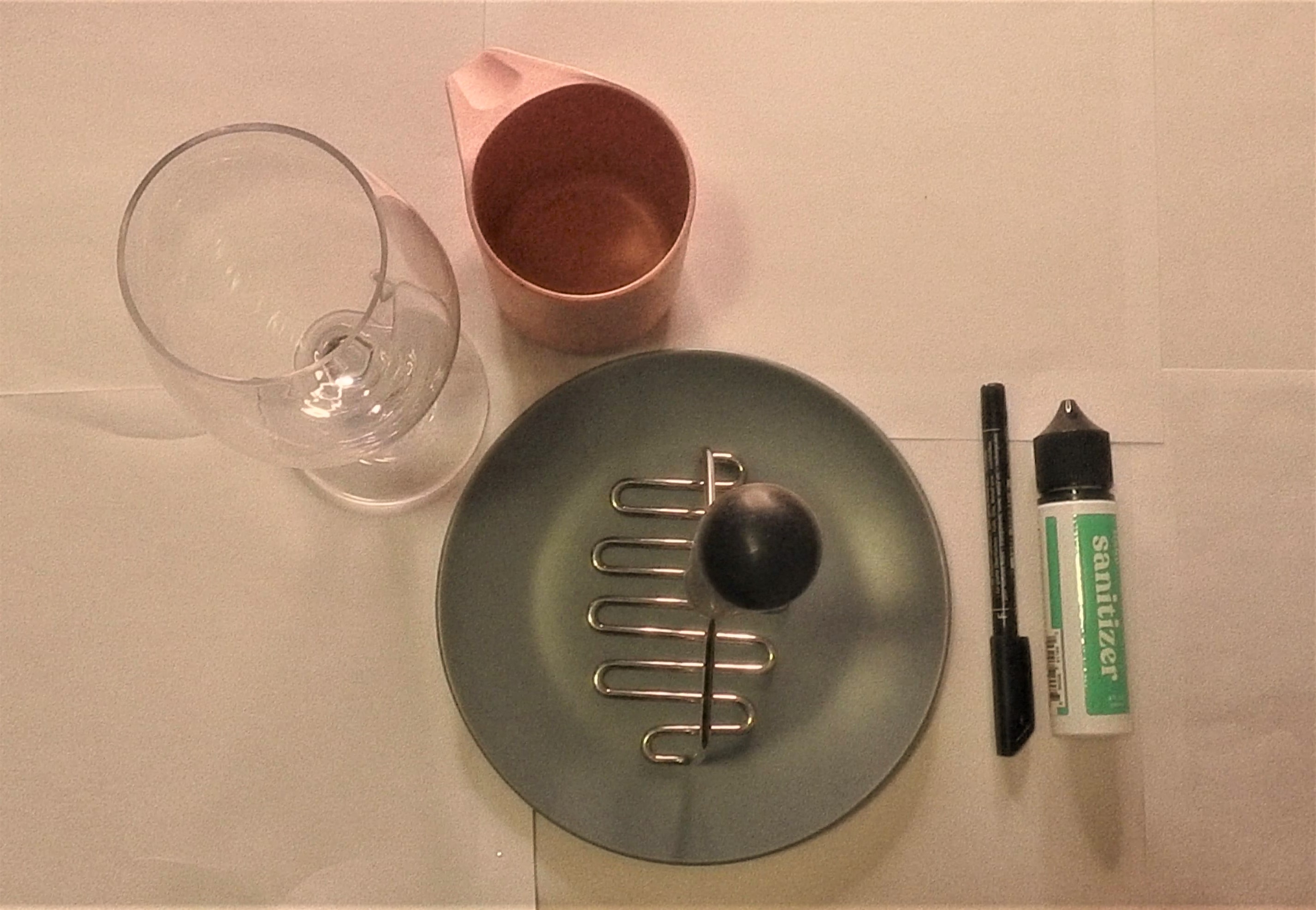}
    \caption{A top-down view of the various domestic objects used to demonstrate variable geometry and compliance: plate, mug, wine glass, pen, sanitizer bottle, and masher.}
    \label{fig:target_objects}
\end{figure}

The various target demonstrations were chosen to capture various aspects of the challenge of manipulating in homes. Fig. \ref{fig:target_objects} depicts the set of domestic objects that were used for this study.

The commanded bubble pressures were in the range of 1010~hPa - 1090~hPa. The demonstrated range of pressure is a function of the elasticity of the membrane and was chosen from empirical observations on safe levels of inflation. The setup as described in Sec. \ref{sec:pressure_regulation} presents a large time period between the inflation and deflation times due to the absence of a method to suck out air from the bubbles, however this is mitigated by an increased wait delay for deflation.

The setup, as well as the various experimental tasks, can also be seen in greater detail in the video accompanying this paper.

\subsection{Geometry variation and contact patch size}
\begin{figure}[ht]
    \centering 
    \includegraphics[width=0.975\columnwidth]{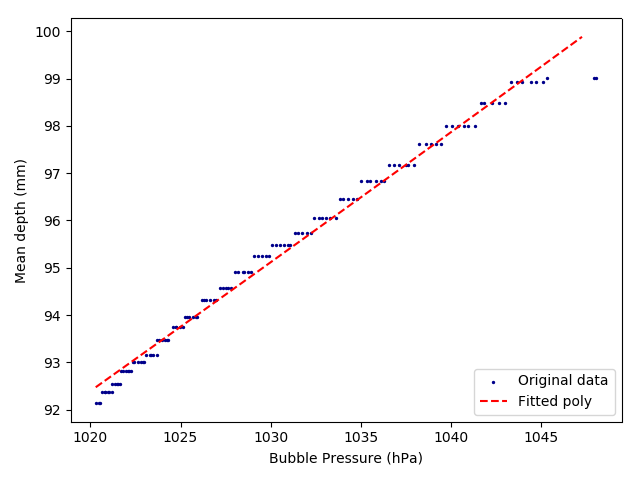}
    \caption{Quantifying the geometry change due to pressure regulation: the change in the mean depth over the entire tactile image due to change in regulated pressure when not grasping. This indicates that the pressure increase is a reasonably good predictor of geometry increase of the entire bubble when not in contact with objects.}
    \label{fig:geometry_pressure_variation}
\end{figure}

\begin{figure}[ht]
    \centering 
    \includegraphics[width=0.975\columnwidth]{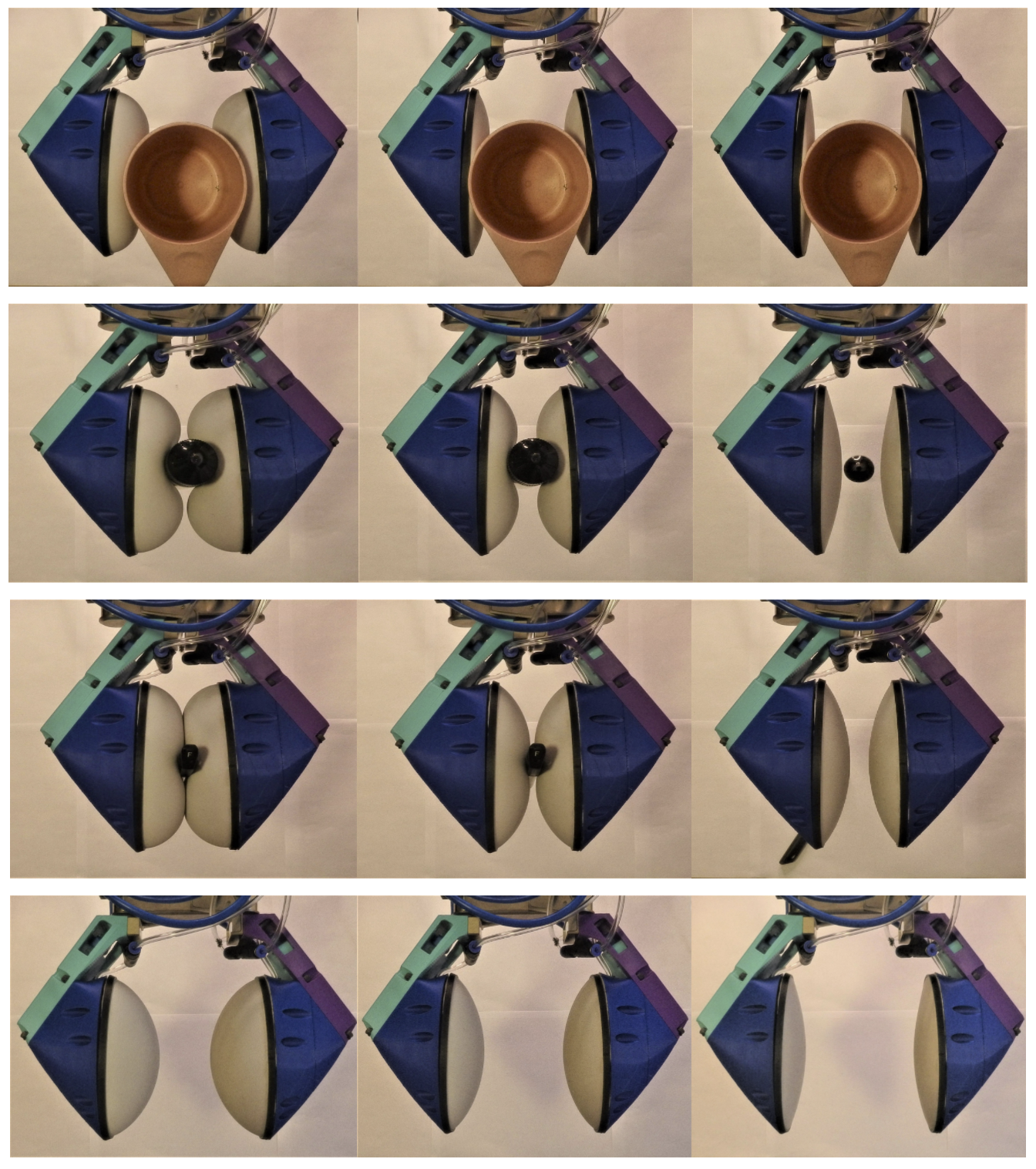}
    \caption{The variation of contact patch size with different pressure levels for a 25~N grasp on various objects: large radius object - mug (topmost row),  medium radius object - sanitizer bottle (second row), small radius object - pen (third row) and for comparison, the no object case (bottom row). The commanded bubble pressures were 1070~hPa (left column), 1050~hPa (middle column), and 1020~hPa (right column). Under low pressures, the narrow form factor objects are no longer able to be grasped. From the geometry of the bubble while grasping, it can be seen that contacted area gradually increases with pressure until saturation.}
    \label{fig:compliance_and_geometry}
\end{figure}

In analyzing the variation of the bubble geometry and the resulting change in contact patch size, it is important to note the elliptical form factor of the base geometry of the bubble resulting in ellipsoidal inflated bubble geometry. We quantified the change in geometry due to pressure variation by measuring the average depth over the entire tactile depth image (as obtained from the internal ToF camera) under a no-contact condition. From the results that can be seen in Fig. \ref{fig:geometry_pressure_variation}, the resulting change in mean depth with a change in pressure from 1050~hPa to 1010~hPa is near linear (the second order coefficient of a polynomial fit was $1.3\times 10^{-4} mm/{hPa^2}$). The consequence of this geometry change on a grasped object can be seen qualitatively depicted in the results in  in Fig. \ref{fig:compliance_and_geometry}; the bubbles expand as internal pressure increases and this results in an increasing contact patch area - this patch area change, along with membrane tension, also influences the stiffness of the obtained grasp as well as the contact friction. Each of these grasps utilized a grasp force of 25~N and the grasp width is left unspecified.

\begin{figure}[ht]
    \centering 
    \includegraphics[width=0.975\columnwidth]{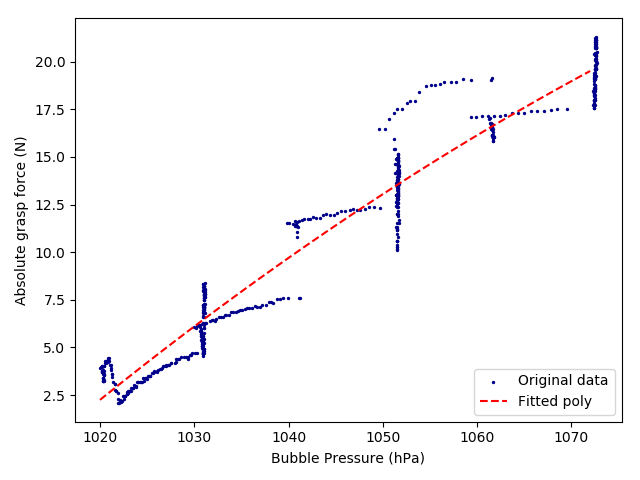}
    \caption{Quantifying the stiffness change due to pressure regulation: the change in the absolute measured grasp force (N) when maintaining grasp on a cylindrical object due to changes in regulated pressure. Since the grasp is  width remains near constant (measured standard deviation of $6.8 \times 10^{-4} mm$), the measured force change indicates that the stiffness increases nearly linearly with the increase in pressure.}
    \label{fig:stiffness_pressure_variation}
\end{figure}

To quantify the relationship between pressure change and grasp stiffness, we measured the change in grasp force as reported by the Schunk gripper under various bubble pressures. For consistency in procedure, the robot was commanded to execute a grasp of 25N on a cylindrical object of 44~mm in diameter and only a single bubble's pressure was varied over a range of 1070~hPa to a completely deflated state of 1010~hPa. The results of this experiment as seen in Fig. \ref{fig:stiffness_pressure_variation} indicate a near linear relationship between measured grasp force and pressure variation (the second order coefficient of a polynomial fit was $-0.00127 N/{hPa^2}$). Since the grasp was of near constant width (measured with a standard deviation of $6.8 \times 10^{-4}mm$), this result indicates that the stiffness increase in grasps on the tested cylindrical object geometry increases linearly with bubble pressure.

The application examples for geometry and stiffness variation include that of plate picking, mug/wine glass picking and manipulating a tool such as a masher. In each of these cases, the appropriate grasp pressure was found through experimental testing. The results of these tasks can be seen in the sensor visualizations in Fig. \ref{fig:Shear_sensing} and in the accompanying video submission. These demonstrations clearly show the utility of the proposed hardware and framework. 

\subsection{Contact patch and shear estimation}

As described in Sec. \ref{sec:shear_estimation}, the variable-geometry shear estimation method results in computing shear solely due to interaction with the object; the computation of the appropriate depth mask depends on the current inflation level. In practice, this was accomplished by simply recapturing a reference depth image in order to compute a mask prior to making contact - i.e. inflate, capture depth mask image, proceed with grasp. While the obtained shear was analyzed qualitatively, the quantitative measurements are outside the scope of this paper. 

\begin{figure}[t]
    \centering 
\begin{subfigure}{0.45\textwidth}
  \includegraphics[width=\linewidth]{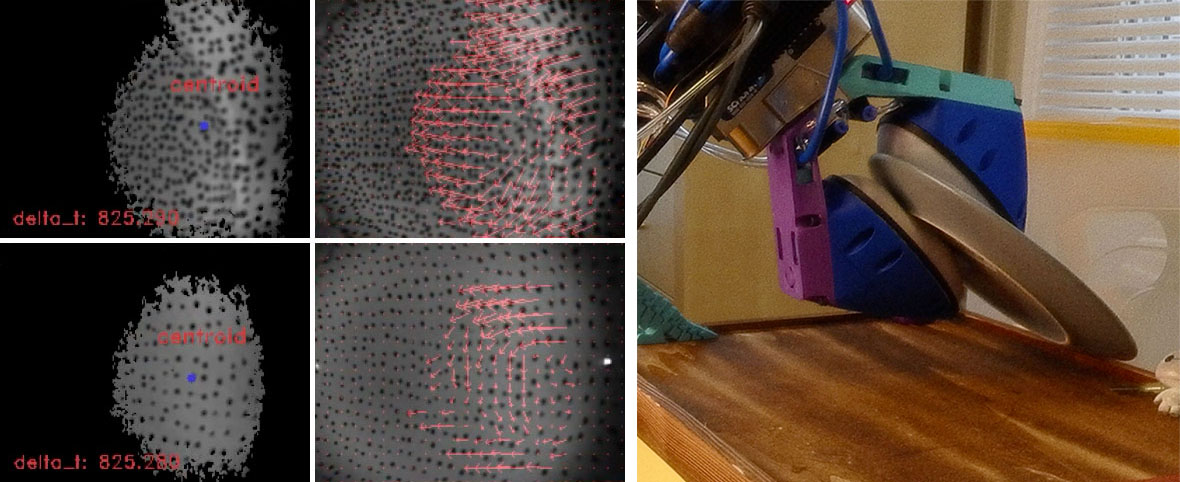}
  \label{fig:Shear_sensing:plate}
\end{subfigure}\hfil 

\vspace*{-1mm}

\begin{subfigure}{0.45\textwidth}
  \includegraphics[width=\linewidth]{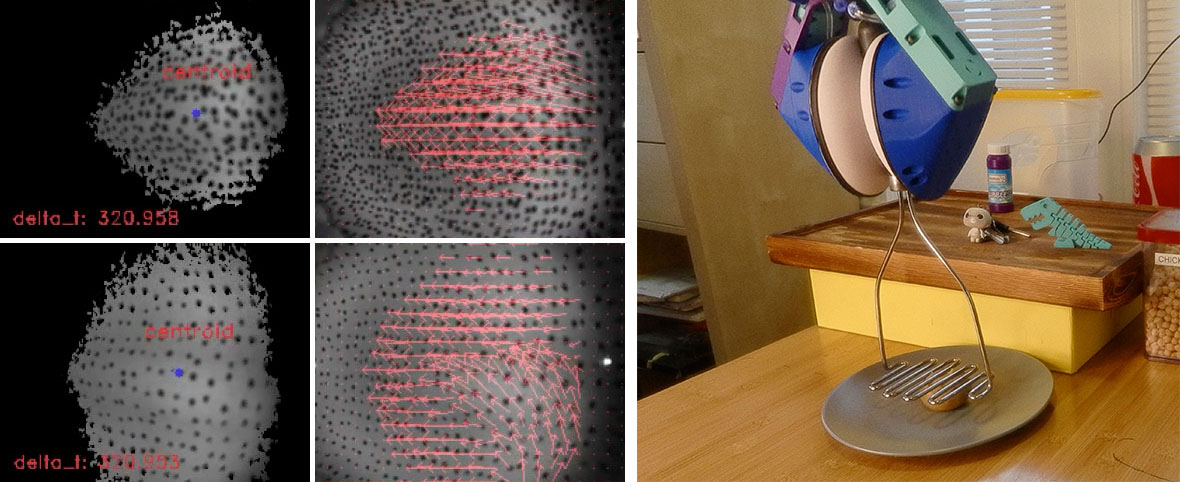}
  \label{fig:Shear_sensing:mash}
\end{subfigure}\hfil 

\vspace*{-1mm}

\begin{subfigure}{0.45\textwidth}
  \includegraphics[width=\linewidth]{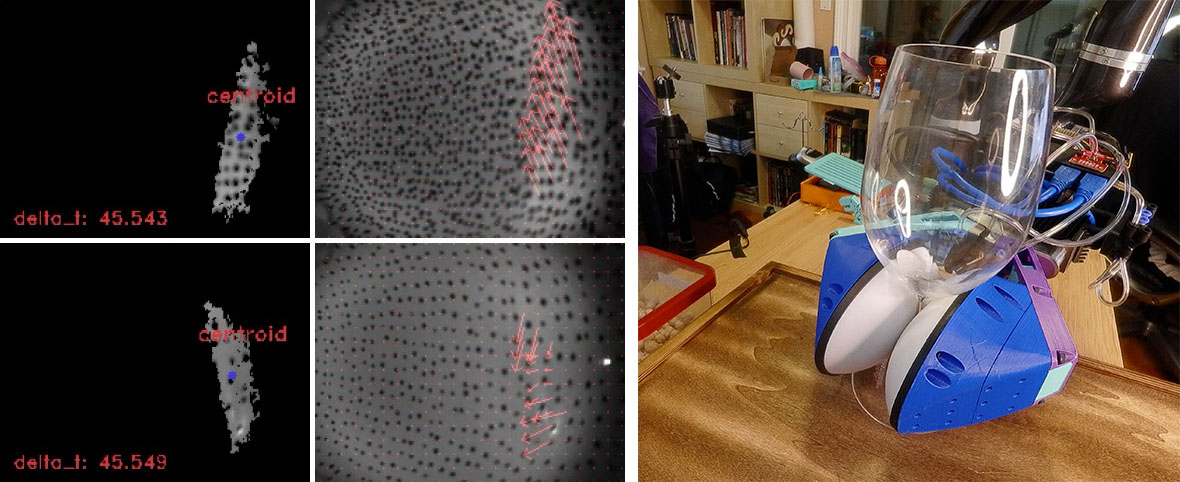}
  \label{fig:Shear_sensing:wine}
\end{subfigure}\hfil 
\caption{Sensor visualizations depicting bubble geometry-independent shear sensing for three different kinds of grasps: Contact-patch masked image and its centroid (left column), shear-induced displacement computed over the contact patch (middle column), and an external view of the obtained grasp (right column). The measured shear in each case is dependent on the grasp geometry and and takes the inflation state into account.}
\label{fig:Shear_sensing}
\end{figure}

Fig. \ref{fig:Shear_sensing} depicts the computed shear at critical phases of grasping and manipulation of a set of the target objects in their respective tasks. The mask closely captures the contact patches while the computed shear closely correlates with the expected force directions.

\section{Conclusions and Discussions}
\label{sec:conclusions}
In this paper we presented a framework for variable compliance and geometry through active pressure regulation of the \Softbubble gripper. Our proposed system uses an active pump-based pressure inflation/deflation system that is small enough to be included on an end effector. We present a method to compute a contact patch along the gripper and estimate shear based on optical flow.
This system was tested in a variety of home robotics related tasks with promising results in the grasp variety and quality. We have demonstrated the system is capable of deflating to widen the gripper opening for larger objects while being able to grasp smaller objects by inflating. Adjusting the pressure within the bubble allows for the contact patch area as well as stiffness to be changed according to desired compliance of the task. Through experiments, we show that there are near-linear relationships between bubble geometry and grasp stiffness with the bubble pressure. 

The ability to change the stiffness of the gripper interface allows an otherwise highly compliant gripper to function in more forceful applications such as the demonstrated potato mashing task. Varying the contact patch through these means also allows us to change the friction between the gripper and the object, which could enable behaviors that involve controlled slip.

Our proposed method provides pathway toward a highly capable gripper that offers compliance that is tunable to the application based on the ability to change geometry, friction, or force transmission requirements. We believe that future controllers, both model- and learning-based, will better cope with real world situations if they are able to change gripper geometry and compliance on the fly.

\bibliographystyle{IEEEtran}
\bibliography{IEEEabrv,references} 

\begin{thebibliography}{10}
\providecommand{\url}[1]{#1}
\csname url@samestyle\endcsname
\providecommand{\newblock}{\relax}
\providecommand{\bibinfo}[2]{#2}
\providecommand{\BIBentrySTDinterwordspacing}{\spaceskip=0pt\relax}
\providecommand{\BIBentryALTinterwordstretchfactor}{4}
\providecommand{\BIBentryALTinterwordspacing}{\spaceskip=\fontdimen2\font plus
\BIBentryALTinterwordstretchfactor\fontdimen3\font minus
  \fontdimen4\font\relax}
\providecommand{\BIBforeignlanguage}[2]{{%
\expandafter\ifx\csname l@#1\endcsname\relax
\typeout{** WARNING: IEEEtran.bst: No hyphenation pattern has been}%
\typeout{** loaded for the language `#1'. Using the pattern for}%
\typeout{** the default language instead.}%
\else
\language=\csname l@#1\endcsname
\fi
#2}}
\providecommand{\BIBdecl}{\relax}
\BIBdecl

\bibitem{Hughes2016}
\BIBentryALTinterwordspacing
J.~Hughes, U.~Culha, F.~Giardina, F.~Guenther, A.~Rosendo, and F.~Iida, ``Soft
  manipulators and grippers: A review,'' \emph{Frontiers in Robotics and AI},
  vol.~3, p.~69, 2016. [Online]. Available:
  \url{https://www.frontiersin.org/article/10.3389/frobt.2016.00069}
\BIBentrySTDinterwordspacing

\bibitem{shintake2018soft}
J.~Shintake, V.~Cacucciolo, D.~Floreano, and H.~Shea, ``Soft robotic
  grippers,'' \emph{Advanced Materials}, vol.~30, no.~29, p. 1707035, 2018.

\bibitem{wolf2015variable}
S.~Wolf, G.~Grioli, O.~Eiberger, W.~Friedl, M.~Grebenstein, H.~H{\"o}ppner,
  E.~Burdet, D.~G. Caldwell, R.~Carloni, M.~G. Catalano \emph{et~al.},
  ``Variable stiffness actuators: Review on design and components,''
  \emph{IEEE/ASME transactions on mechatronics}, vol.~21, no.~5, pp.
  2418--2430, 2015.

\bibitem{ruggiero2018nonprehensile}
F.~Ruggiero, V.~Lippiello, and B.~Siciliano, ``Nonprehensile dynamic
  manipulation: A survey,'' \emph{IEEE Robotics and Automation Letters},
  vol.~3, no.~3, pp. 1711--1718, 2018.

\bibitem{kuppuswamy2019fast}
N.~Kuppuswamy, A.~Castro, C.~Phillips-Grafflin, A.~Alspach, and R.~Tedrake,
  ``Fast model-based contact patch and pose estimation for highly deformable
  dense-geometry tactile sensors,'' \emph{IEEE Robotics and Automation
  Letters}, 2019.

\bibitem{Kuppuswamy2020}
N.~{Kuppuswamy}, A.~{Alspach}, A.~{Uttamchandani}, S.~{Creasey}, T.~{Ikeda},
  and R.~{Tedrake}, ``{Soft-Bubble grippers for robust and perceptive
  manipulation},'' \emph{arXiv e-prints}, p. arXiv:2004.03691, Apr. 2020.

\bibitem{Alspach2019}
A.~Alspach, K.~Hashimoto, N.~Kuppuswamy, and R.~Tedrake, ``Soft-bubble: A
  highly compliant dense geometry tactile sensor for robot manipulation,'' in
  \emph{2019 2nd IEEE International Conference on Soft Robotics
  (RoboSoft)}.\hskip 1em plus 0.5em minus 0.4em\relax IEEE, 2019, pp. 597--604.

\bibitem{schmidt1978flexible}
I.~Schmidt, ``Flexible moulding jaws for grippers,'' \emph{Industrial Robot: An
  International Journal}, 1978.

\bibitem{Chang2019}
\BIBentryALTinterwordspacing
C.-M. Chang, L.~Gerez, N.~Elangovan, A.~Zisimatos, and M.~Liarokapis, ``On
  alternative uses of structural compliance for the development of adaptive
  robot grippers and hands,'' \emph{Frontiers in Neurorobotics}, vol.~13,
  p.~91, 2019. [Online]. Available:
  \url{https://www.frontiersin.org/article/10.3389/fnbot.2019.00091}
\BIBentrySTDinterwordspacing

\bibitem{ozawa2017grasp}
R.~Ozawa and K.~Tahara, ``Grasp and dexterous manipulation of multi-fingered
  robotic hands: a review from a control view point,'' \emph{Advanced
  Robotics}, vol.~31, no. 19-20, pp. 1030--1050, 2017.

\bibitem{schmitt2018}
\BIBentryALTinterwordspacing
F.~Schmitt, O.~Piccin, L.~Barbé, and B.~Bayle, ``Soft robots manufacturing: A
  review,'' \emph{Frontiers in Robotics and AI}, vol.~5, p.~84, 2018. [Online].
  Available: \url{https://www.frontiersin.org/article/10.3389/frobt.2018.00084}
\BIBentrySTDinterwordspacing

\bibitem{Kim2015}
J.~Kim, A.~Alspach, and K.~Yamane, ``3d printed soft skin for safe human-robot
  interaction,'' in \emph{IROS}.\hskip 1em plus 0.5em minus 0.4em\relax IEEE,
  2015, pp. 2419--2425.

\bibitem{Alspach2015}
A.~Alspach, J.~Kim, and K.~Yamane, ``Design of a soft upper body robot for
  physical human-robot interaction,'' in \emph{Humanoids}, Nov 2015, pp.
  290--296.

\bibitem{Yamaguchi2019}
\BIBentryALTinterwordspacing
A.~Yamaguchi and C.~G. Atkeson, ``Recent progress in tactile sensing and
  sensors for robotic manipulation: can we turn tactile sensing into vision?''
  \emph{Advanced Robotics}, vol.~33, no.~14, pp. 661--673, 2019. [Online].
  Available: \url{https://doi.org/10.1080/01691864.2019.1632222}
\BIBentrySTDinterwordspacing

\bibitem{arachchige2020novel}
D.~D. Arachchige, Y.~Chen, I.~D. Walker, and I.~S. Godage, ``A novel variable
  stiffness soft robotic gripper,'' 2020.

\bibitem{paoletti2017grasping}
P.~Paoletti, G.~Jones, and L.~Mahadevan, ``Grasping with a soft glove:
  intrinsic impedance control in pneumatic actuators,'' \emph{Journal of the
  Royal Society Interface}, vol.~14, no. 128, p. 20160867, 2017.

\bibitem{adami2019board}
M.~Adami and A.~Seibel, ``On-board pneumatic pressure generation methods for
  soft robotics applications,'' in \emph{Actuators}, vol.~8, no.~1.\hskip 1em
  plus 0.5em minus 0.4em\relax Multidisciplinary Digital Publishing Institute,
  2019, p.~2.

\bibitem{Best2020}
\BIBentryALTinterwordspacing
C.~M. Best, L.~Rupert, and M.~D. Killpack, ``Comparing model-based control
  methods for simultaneous stiffness and position control of inflatable soft
  robots,'' \emph{The International Journal of Robotics Research}, vol.~0,
  no.~0, p. 0278364920911960, 0. [Online]. Available:
  \url{https://doi.org/10.1177/0278364920911960}
\BIBentrySTDinterwordspacing

\bibitem{Farneback2003}
G.~Farneback, ``Two-frame motion estimation based on polynomial expansion.'' in
  \emph{SCIA}, 2003.

\bibitem{Yuan2015Shear}
W.~{Yuan}, R.~{Li}, M.~A. {Srinivasan}, and E.~H. {Adelson}, ``Measurement of
  shear and slip with a gelsight tactile sensor,'' in \emph{2015 IEEE
  International Conference on Robotics and Automation (ICRA)}, 2015, pp.
  304--311.

\bibitem{Taylor2005_fem_implementation}
R.~L. Taylor, E.~O{\~n}ate, and P.-A. Ubach, ``Finite element analysis of
  membrane structures,'' in \emph{Textile Composites and Inflatable
  Structures}.\hskip 1em plus 0.5em minus 0.4em\relax Springer, 2005, pp.
  47--68.

\bibitem{Shimonomura2019}
K.~Shimonomura, ``Tactile image sensors employing camera: A review,''
  \emph{Sensors}, vol.~19, no.~18, p. 3933, 2019.

\bibitem{yuan2015measurement}
W.~Yuan, R.~Li, M.~A. Srinivasan, and E.~H. Adelson, ``Measurement of shear and
  slip with a {GelSight} tactile sensor,'' in \emph{ICRA}.\hskip 1em plus 0.5em
  minus 0.4em\relax IEEE, 2015, pp. 304--311.

\bibitem{zhang2019}
\BIBentryALTinterwordspacing
Y.~Zhang, Z.~Kan, Y.~Yang, Y.~A. Tse, and M.~Y. Wang, ``Effective estimation of
  contact force and torque for vision-based tactile sensor with helmholtz-hodge
  decomposition,'' \emph{CoRR}, vol. abs/1906.09460, 2019. [Online]. Available:
  \url{http://arxiv.org/abs/1906.09460}
\BIBentrySTDinterwordspacing

\bibitem{bib:drake}
\BIBentryALTinterwordspacing
R.~Tedrake and the Drake Development~Team, ``Drake: Model-based design and
  verification for robotics,'' 2019. [Online]. Available:
  \url{https://drake.mit.edu}
\BIBentrySTDinterwordspacing

\end{thebibliography}
\end{document}